\documentclass[letterpaper, 10 pt, conference]{ieeeconf}

\IEEEoverridecommandlockouts                              

\usepackage[nolist,nohyperlinks]{acronym}

\usepackage{cite}

\usepackage{xcolor}

\newcommand\submapmath[1]{\mathbb{#1}}

\newcommand{\mat}[1]{\mathbf{#1}}

\usepackage{amsfonts}
\usepackage{amsmath}

\usepackage{graphicx}
\usepackage[font=footnotesize,labelfont=bf]{caption}
\usepackage{subcaption}

\usepackage{booktabs}

\usepackage{url}

\renewcommand\vec{\mathbf}

\usepackage{pgfplots}

\usepackage{textcomp}

\pgfplotsset{compat=newest}
\newlength\figureheight
\newlength\figurewidth

\makeatletter
    \@ifdefinable{\submap}{\def\submap/{submap}}
    \@ifdefinable{\Submap}{\def\Submap/{Submap}}
    \@ifdefinable{\rgbd}{\def\rgbd/{RGB-D}}
\makeatother

\title{\LARGE \bf
Freetures:\\Localization in Signed Distance Function Maps
}
\author{Alexander Millane\textsuperscript{$\ddagger$}, Helen Oleynikova\textsuperscript{$\dagger$}, Christian Lanegger\textsuperscript{$\ddagger$}, Jeff Delmerico\textsuperscript{$\dagger$} \\ Juan Nieto\textsuperscript{$\dagger$}, Roland Siegwart\textsuperscript{$\ddagger$}, Marc Pollefeys\textsuperscript{$\dagger$}, C\'{e}sar Cadena\textsuperscript{$\ddagger$}\\
 \textsuperscript{$\ddagger$}Autonomous Systems Lab, ETH  Z{\"u}rich,
 \textsuperscript{$\dagger$}Microsoft Mixed Reality and AI Zurich Lab
\thanks{This research was funded by the National Center of Competence in Research (NCCR) Robotics through the Swiss National Science Foundation and Microsoft.}
}

\begin{acronym}
\acro{SLAM}{Simultaneous Localization And Mapping}
\acro{SDF}{Signed Distance Function}
\acro{TSDF}{Truncated Signed Distance Function}
\acro{ESDF}{Euclidean Signed Distance Function}
\acro{DoH}{Determinant of Hessian}
\acro{MAV}{Micro Aerial Vehicle}
\acro{IMU}{Inertial Measurement Unit}
\acro{FPFH}{Fast Point Feature Histograms}
\acro{RANSAC}{Random Sample Consensus}
\acro{ICP}{Iterative Closest Point}
\acro{NARF}{Normal-Aligned Radial Features}
\acro{BoW}{Bag of Words}
\acro{LRF}{Local Reference Frame}
\acro{HOG}{Histogram of Oriented Gradients}
\acro{SIFT}{Scale-Invariant Features Transform}
\acro{FLANN}{Fast Library for Approximate Nearest Neighbors}
\acro{SHOT}{Signature of Histograms of OrienTations}
\acro{RTK}{Real-Time Kinematic}
\end{acronym}

\begin{document}
\maketitle


\begin{abstract}
Localization of a robotic system within a previously mapped environment is important for reducing estimation drift and for reusing previously built maps. Existing techniques for geometry-based localization have focused on the description of local surface geometry, usually using pointclouds as the underlying representation. We propose a system for geometry-based localization that extracts features directly from an implicit surface representation: the \ac{SDF}. The \ac{SDF} varies continuously through space, which allows the proposed system to extract and utilize features describing both surfaces and free-space. Through evaluations on public datasets, we demonstrate the flexibility of this approach, and show an increase in localization performance over state-of-the-art handcrafted surfaces-only descriptors. We achieve an average improvement of \texttildelow12\% on an \rgbd/ dataset and \texttildelow18\% on a LiDAR-based dataset. Finally, we demonstrate our system for localizing a LiDAR-equipped \ac{MAV} within a previously built map of a search and rescue training ground.
\end{abstract}

\section{Introduction}

Localization is a key capability for robotics systems. During exploratory motion, localization is critical to reducing accumulated estimation drift. During deployment in previously visited environments, localization allows for reuse of existing maps~\cite{cadena2016past}. Techniques leveraging 3D data are particularly important for the robots automating our homes, factories, and cities, in part because these robots frequently make use of 3D sensing, for example LiDAR or depth-cameras, but also because of the complementary properties of 3D sensing to images, such as illumination invariance and geometric accuracy. 

Existing approaches to 3D localization are typically based on extraction and description of features from pointcloud data, with several successful approaches proposed~\cite{bosse2013place, steder2011place, dube2020segmap, rusu2009fast, choi2015robust, salti2014shot}. Dense representations based on \acp{SDF} have also become increasingly popular, primarily for their success in fusing noisy \rgbd/ data, but also for fusing LiDAR data~\cite{reijgwart2019voxgraph}, for generating consistent maps~\cite{fioraio2015large}, and for robotic path-planning~\cite{oleynikova2017voxblox}. \acp{SDF} have proven themselves a useful alternative to pointcloud-based representations in many contexts, but are relatively unexplored for use in localization. In this work we investigate such an approach.

We introduce a method for geometry-based localization in dense 3D maps. We aim to locate an agent within a previously observed scene, which is represented as a collection of \submap/s. Surface geometry within each \submap/ is stored as an implicit surface, the \ac{SDF}. We extract local features \textit{directly} on the \ac{SDF}, an idea first introduced in 2D in our previous work~\cite{millane2019freespace}, which is further developed, and applied to 3D data, here. The approach allows for characterization of both surface and free-space geometry, since the \ac{SDF} treats both of these spaces equally. Our hypothesis is that by extending the region of description beyond surface boundaries alone, localization performance can be improved. We evaluate this hypothesis by testing the efficacy of our system for localization on public and self-collected \rgbd/ and LiDAR datasets.

\begin{figure}
    \centering
    \includegraphics[width=0.95\columnwidth]{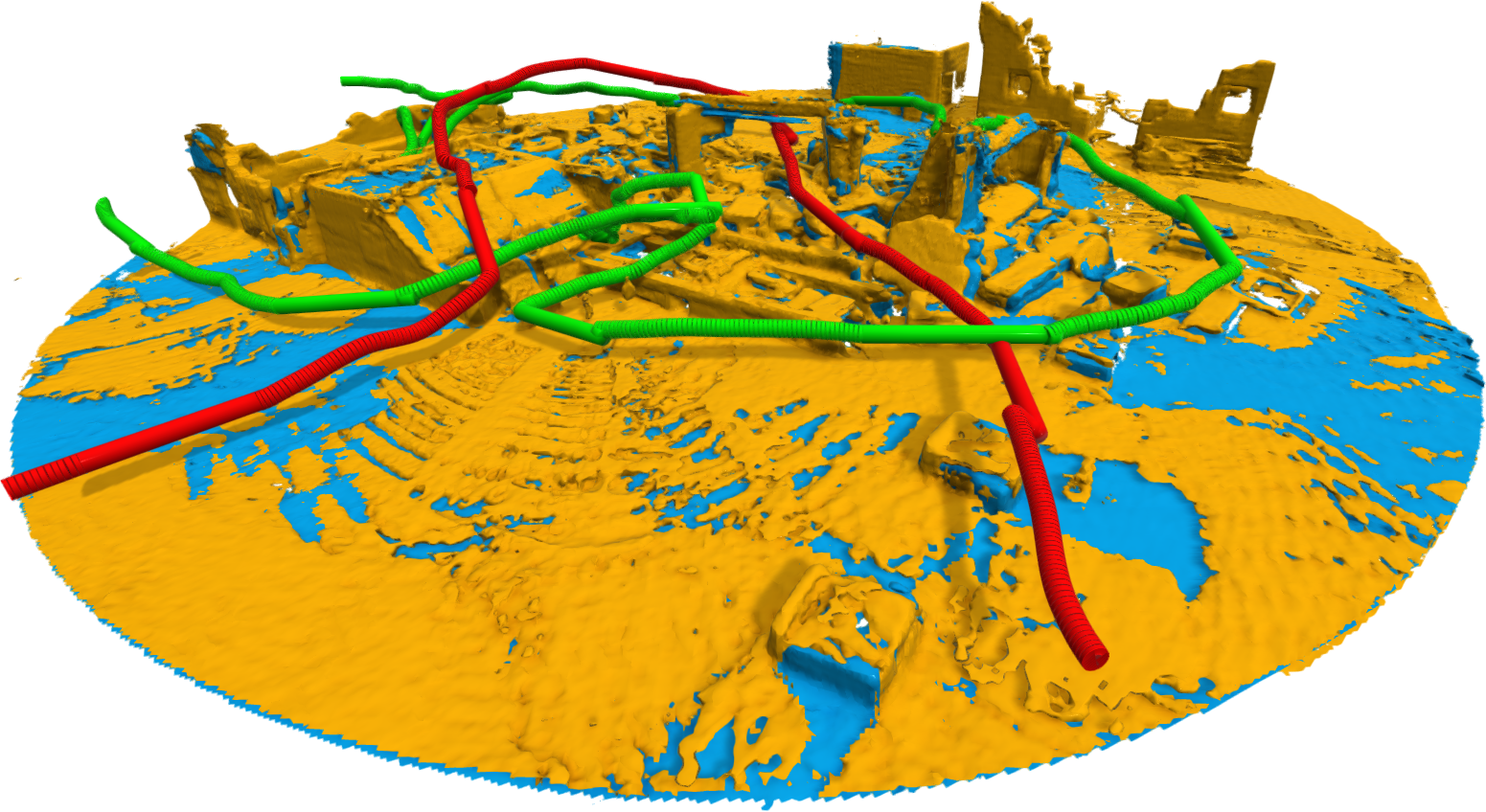}
    \caption{Two maps of rubble from a collapsed building, produced by a LiDAR equipped \ac{MAV} during two flights through a search and rescue training ground (estimated \ac{MAV} trajectories are shown in red and green). The query map (yellow) is registered against the target map (blue) using the proposed localization system. See Sec.~\ref{sec:results_arche} for details.}
    \label{fig:teaser}
    \vspace{-5mm}
\end{figure}

In summary, the contributions of this paper are:
\begin{itemize}
    \item An investigation into the use of distance functions explicitly for the purpose of localization in 3D
    \item A keypoint detection and description approach for characterization of local \ac{SDF} geometry
    \item An open-source implementation of the proposed system\footnote{https://github.com/alexmillane/freetures}
\end{itemize}

\section{Related Work}
In this section, we give a brief review of SDF-based mapping methods and an overview of a number of different localization methods. We will compare these to our proposed SDF-based localization method.

\subsection{Localization in \ac{SDF}-based Mapping Systems}
Implicit surfaces have become a popular means for representing geometry in mapping systems. \acfp{TSDF}, 
popularized by KinectFusion~\cite{izadi2011kinectfusion}, have proven particularly successful for fusing noisy, high-rate data from commodity depth-cameras. Several authors have since improved on these seminal works, enabling application of \ac{TSDF}s to larger environments~\cite{niessner2013real}, for maintaining global-consistency \cite{fioraio2015large}, for fusing LiDAR data~\cite{reijgwart2019voxgraph}, and for use in robotic path-planning~\cite{oleynikova2017voxblox}. Distance function-based representations are, at present, one of the most widely used for dense mapping.

In order to maintain map consistency, several works have explored localization in the context of \ac{SDF}-based \ac{SLAM}. The most common approach is to detect image features in \rgbd/ frames and leverage successful localization systems developed for visual \ac{SLAM}~\cite{dai2017bundlefusion, millane2018c, reijgwart2019voxgraph}. 
BundleFusion~\cite{dai2017bundlefusion}, for example, relies on \ac{SIFT} image feature matching~\cite{lowe1999object}. In a more specialized approach, Glocker \textit{et al}.~\cite{glocker2014real} suggest a feature designed specifically for \rgbd/ frames. 

The use of image features has several disadvantages in the context of this work. Firstly, observation data produced by some common sensors do not contain image data, for example time-of-flight based cameras or LiDAR (see Sec.~\ref{sec:results_deutsches}). Secondly, large viewpoint or illumination differences during place-revisiting can render image-based localization challenging~\cite{lowry2015visual}. 3D data in this context has complimentary properties. Lastly, performing localization on the \ac{SDF}-based map allows for a natural means of localizing between differing sensor modalities. With these motivating reasons, we turn our attention to place-recognition based on 3D data alone.

\subsection{Geometry-based Localization}

Geometry-based localization has received considerable research focus. 
One common approach is to match current observations to an existing map using the same techniques used to generate constraints between successive poses~\cite{behley2018rss, hess2016real}. The primary challenge with this approach is that registration techniques converge to local-minima if not initialized sufficiently close to the globally-optimal solution. Registration is therefore typically attempted with several initializations, with the expectation that one will converge to the correct solution~\cite{behley2018rss}. Google's Cartographer System~\cite{hess2016real} improves this approach by employing a branch and bound technique to efficiently prune candidates early in the matching process, effectively increasing the number of proposals that can be tested. In general, however, this class of techniques requires an accurate pose prior for localization to function. 
This limits their efficacy to small environments or to trajectories without long periods of exploratory motion. In contrast, the method presented in this paper is intended for \textit{global}-localization. 

Local geometric features and robust registration algorithms such as \ac{RANSAC} enable pose estimation independent of priors. Several features for this purpose have been suggested. Bosse and Zlot~\cite{bosse2013place} extract 3D Gestalt descriptors from pointcloud data, and use a voting system to determine localization candidates. Steder \textit{et  al.}~\cite{steder2011place} suggest to extract \ac{NARF} features from pointclouds, and combine this with a \ac{BoW}-based approach for global correspondence search. Segmatch~\cite{dube2020segmap} uses segments, contiguous elements in the scene, as the basis for description and matching. The authors of~\cite{choi2015robust} generate a collection of scene fragments by accumulating consecutive \rgbd/ frames, and perform registration using a popular pointcloud-based descriptor, \ac{FPFH}~\cite{rusu2009fast}. Gawel \textit{et al.}~\cite{gawel20173d} present an evaluation of several of these features for localization.

All of these approaches to 3D localization utilize pointclouds as the basis for place description, and are thus limited to capturing the geometry of surfaces only. Our previous work~\cite{millane2019freespace} introduced \textit{freespace features} for geometry-based localization in the context of 2D LiDAR-based mapping. We proposed extracting features \emph{directly} on the distance function representation of the scene. The results showed that this approach gave significant performance increases over a baseline 2D pointcloud-based descriptor, which we showed was primarily due to the utilization of features extracted in free-space. In the present proposal we continue this investigation, first by extending the approach to 3D and addressing the additional challenges this presents, such as determination of \acp{LRF}. Secondly, we introduce further utilization of the \ac{SDF} for localization, by integrating an \ac{SDF}-based fitness measure into the robust registration pipeline, which we found to further increase performance.

\section{Preliminaries}
\label{sec:preliminaries}

In this section we describe the notation used throughout this paper. Coordinate frames are upper-case letters. Vectors are denoted as bold lower case letters, e.g. $\vec{p}_A \in \mathbb{R}^3$, a 3-dimensional vector expressed in reference frame $A$. Matrices are bold upper-case letters, e.g. $\mat{A}$. We denote the rigid transformation between coordinate frames $A$ and $B$ as $\mat{T}_{AB} \in \text{SE}(3)$, such that $\vec{p}_A = \mat{T}_{AB} \vec{p}_B$. Central to our approach is the \ac{SDF} for representing geometry. As is common in modern reconstruction systems, this function is stored as a spatially sparse set of samples over a voxel grid, where the sparsity comes from the fact that only observed voxels are stored. We denote the space of observed voxel indices as $\Psi \subset \mathbb{Z}^3$. The \ac{SDF} is then described by two functions, $\Phi: \Psi \to \mathbb{R}$ and $\omega: \Psi \to \mathbb{R}$, which respectively map points in observed space to the signed distance to the nearest observed surface, and to a weighting/confidence measure. 

\section{Problem Statement}

\begin{figure*}
    \centering
    \includegraphics[width=\textwidth]{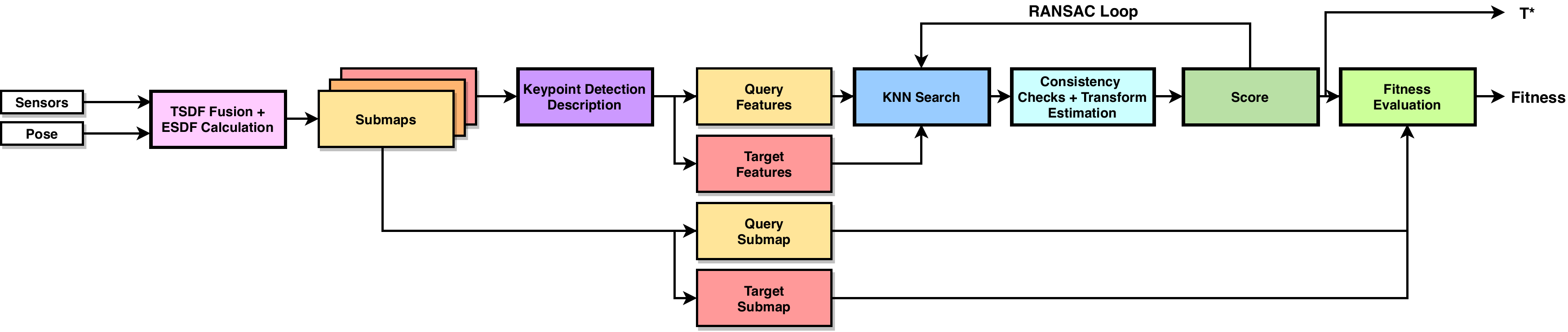}
    \caption{A diagram detailing data flow in our system. The front-end that produces \submap/s is a product of our previous works~\cite{millane2018c, reijgwart2019voxgraph}. The proposed method identifies \submap/s within a collection that describe overlapping regions in the environment. }
    \label{fig:system_diagram}
\end{figure*}

Given a set of \ac{SDF} \submap/s $\{\submapmath{S}_k\}_{k=1}^N$ and associated coordinate frames $\{S_k\}_{k=1}^N$, we consider the problem of finding pairs of \submap/s $(\submapmath{S}_i, \submapmath{S}_j)$ that correspond to the same location in the environment, that is, contain significant overlap in the region of the environment they describe. In addition, we aim to determine $\mat{T}_{S_i S_j} \in \text{SE}(3)$, the rigid transformation describing the \submap/s' relative pose.

\section{Method}

Our approach is described in detail in the following sections, with an overview of the steps shown in Fig. \ref{fig:system_diagram}.

\subsection{\Submap/ Construction}
\label{sec:submap_construction}
The input to the proposed system is a set of \submap/s $\{\submapmath{S}_k\}_{k=1}^N$. For \submap/ construction we make use of several of our previous works. Sensor data is integrated into a \submap/ \ac{TSDF} by ray casting into a spatially-hashed sparse voxel grid (see \cite{oleynikova2017voxblox}). Periodically, we create a new \submap/ by defining a voxel grid, with a coordinate frame co-located and aligned with the sensor pose at the time of \submap/ creation (see \cite{millane2018c}). After completion of a submap, we follow~\cite{oleynikova2017voxblox, reijgwart2019voxgraph} and extract the \ac{ESDF}, as well as a set of points on the zero-level set. For the remainder of this paper, we will refer to the \ac{ESDF} when discussing the \ac{SDF}.

\subsection{Keypoint Detection}
\label{sec:keypoint_detection}
We aim to detect keypoints that can be reliably re-detected during place-revisiting. The \ac{SDF} is smooth by definition and has a gradient magnitude of 1 \textit{almost everywhere}. Typical keypoint detectors, which have high response in areas of large gradient, at image corners for example, do not function well when applied to the \ac{SDF} as a result. Therefore, we detect keypoints in areas of high curvature, as local-extrema of the \ac{DoH} volume, $\delta: \Psi \to \mathbb{R}$. In particular, for a voxel with center indices $\vec{k} \in \Psi$,
\begin{equation}
    \delta(\vec{k}) = \det(\mat{H}(\vec{k})),
    \label{eq:doh_response}
\end{equation}
with $\mat{H}(\vec{k})$ the Hessian,
\begin{equation}
    \mat{H}(\vec{k})
    = 
    \begin{bmatrix}
    h_{xx}(\vec{k}) & h_{xy}(\vec{k}) & h_{xz}(\vec{k})\\
    h_{yx}(\vec{k}) & h_{yy}(\vec{k}) & h_{yz}(\vec{k})\\
    h_{zx}(\vec{k}) & h_{zy}(\vec{k}) & h_{zz}(\vec{k})
    \end{bmatrix},
\end{equation}
where the components are computed through application of Sobel derivative kernels, as well as additional Gaussian blur with tunable variance, set to $\sigma_{\text{grad}}=2\,\text{voxels}$ for the remainder of this work. The first Hessian element $h_{xx}(\vec{k})$, for example, is given by:
\begin{align}
    h_{xx}(\vec{k})
    &=
    (\mat{D}_x * \mat{D}_x * \mat{G} * \Phi)(\vec{k})
    \\
    &=
    (\mat{K} * \Phi)(\vec{k}).
\end{align}
Here, $\mat{D}_x$ is the first-order Sobel derivative kernel in the $x$ direction, $\mat{G}$ a Gaussian kernel, $\Phi$ is the \ac{SDF}, and $*$ the convolution operator. Note that because our data is distributed sparsely over the voxel grid, valid computation of the components of $\mat{H}$ at $\vec{k}$ requires that $\Phi$ be defined over the support of $\mat{K}$ at $\vec{k}$. A voxel is selected as a keypoint if it is an extremum point relative to its immediate neighbours. For \submap/ $\mathbb{S}_i$, the set of keypoint locations $\{\vec{k}_{S_i}^j\}_{j=1}^M$ is passed to the next stages for description.

\subsection{Local Reference Frame Assignment}
\label{sec:local_reference_frame}
To achieve invariance of the proposed descriptor to \submap/ poses, we assign a \ac{LRF} to each keypoint, and perform description in this local frame. Several methods for this task have been suggested (see~\cite{petrelli2011repeatability} for a review), however many of the proposed methods rely on the presence of a strong surface normal. Our keypoints are, in general, not located on surfaces, and as a result these methods are not applicable in this context. We found that an approach based on an Eigenvalue decomposition of the gradient field within the descriptor support to generate the most repeatable frames.


We compute the gradient field of the \submap/, $\vec{g}_\text{raw}: \Psi \to \mathbb{R}^3$, through convolution with the Sobel kernel by reusing intermediate results generated in keypoint detection. For each keypoint we weight the gradient vectors in the descriptor support, a sphere with radius $r_f$, with a Gaussian weighting function of tunable variance, set to $\sigma_{\text{desc}} = r_f$ for the remainder of this work. This gives more importance to gradients close to the descriptor center, a common approach. We construct the structure tensor for a keypoint located at $\vec{k}$ based on voxels within the descriptor support $\Omega$,
\begin{equation}
    \mat{S}_{\Omega}(\vec{k}) = \sum_{\vec{k} \in \Omega} \mat{S}(\vec{k}),
\end{equation}
where
\begin{equation}
    \mat{S}(\vec{k})
    =
    \begin{bmatrix}
        g_x(\vec{k})^2 & g_x(\vec{k}) g_y(\vec{k}) & g_x(\vec{k}) g_z(\vec{k}) \\
        g_y(\vec{k}) g_x(\vec{k}) & g_y(\vec{k})^2 & g_y(\vec{k}) g_z(\vec{k}) \\
        g_z(\vec{k}) g_x(\vec{k}) & g_z(\vec{k}) g_y(\vec{k}) & g_z(\vec{k})^2
    \end{bmatrix},
\end{equation}
where $g_x(\vec{k})$ is the $x$ component of the weighted gradient vector, $\vec{g}$ at $\vec{k}$. The (unsigned) axes of our \ac{LRF} are the Eigenvectors of $\mat{S}_{\Omega}$, $\{\vec{v}_1, \vec{v}_2, \vec{v}_3\}$, sorted in descending order according to their corresponding Eigenvalues. To assign directionality to the axes, we project gradient vectors within the descriptor support onto the Eigenvectors. We set the first axis of the \ac{LRF}, $\vec{a}_1$ as
\begin{equation}
    \vec{a}_1 = 
\begin{cases}
    \vec{v}_1                & \text{if } s\geq k_{\text{axis}}\\
    (\vec{v}_1, -\vec{v}_1)  & \text{if } -k_{\text{axis}} \leq s \leq k_{\text{axis}}\\
    -\vec{v}_1               & \text{if } s \leq -k_{\text{axis}}\\
\end{cases}
\end{equation}
where 
\begin{equation}
    s = \frac{\sum_{\vec{k} \in \Omega} \vec{g}(\vec{k}) \cdot \vec{v}_1}{\sum_{\vec{k} \in \Omega} |\vec{g}(\vec{k}) \cdot \vec{v}_1|},
\end{equation}
$\vec{v}_1$ is the first Eigenvector. We perform this procedure for the 1st and 3rd Eigenvectors, and choose the remaining axis direction to complete a right handed coordinate system. Note that, in the case that $-k_{\text{axis}} \leq s \leq k_{\text{axis}}$ we create multiple \ac{LRF}s, one with each direction of the relevant Eigenvector, and perform the description detailed in the Sec.~\ref{sec:description} for each of the \ac{LRF}s. We set $k_{\text{axis}} = 0.5$ for the remainder of this work. The output of this stage is rotation matrix $\mat{R}_{F_i S} \in \text{SO}(3)$ describing the rotation from the \submap/ frame $S$ to the $i^{\text{th}}$ feature frame $F_i$.

\subsection{Description}
\label{sec:description}
For each keypoint and each associated \ac{LRF}, we compute a description based on a gradient orientation histogram, popularized by the \ac{HOG}~\cite{dalal2005histograms} and \ac{SIFT}~\cite{lowe1999object} image features. For the $i^\text{th}$ feature we rotate the weighted gradients from the \submap/ frame, $S$, to the feature frame, $F_i$,
\begin{equation}
    \vec{g}_{F_i}(\vec{k}) = \mat{R}_{F_i S} \vec{g}_{S}(\vec{k}), \quad \forall \vec{k} \in \Omega \cap \Psi,
\end{equation}
where $\Omega \cap \Psi$ denotes valid voxels in the descriptor support, and $\mat{R}_{F_i S}$ is described by the feature \ac{LRF} computed as described in Sec.~\ref{sec:local_reference_frame}.

We then convert the weighted gradients to spherical coordinates 
and compute a weighted histogram by dividing azimuth $\phi \in [-\pi, \pi]$ and polar $\theta  \in [-\frac{\pi}{2}, \frac{\pi}{2}]$ angles into $n_{div}/180^{\circ}$ divisions. We employ soft binning such that each gradient vector contributes its magnitude to the surrounding bins through bilinear interpolation, as suggested by~\cite{dalal2005histograms}. We normalize the histogram counts by the number of voxels in the descriptor support, and each bin by its solid angle (see~\cite{scovanner20073} for details).

We augment the gradient orientation histogram with two further pieces of information. Firstly, we compute the weighted sum of the \ac{SDF} under the descriptor support
\begin{equation}
    b_{\text{dist}} = \frac{1}{N_{b_\text{dist}}} \sum_{\vec{k} \in \Omega \cap \Psi} k_{\text{gauss}}(\vec{k}) \Phi(\vec{k}).
\end{equation}
Where $k_{\text{gauss}}$ is the same central weighting function used to weight the gradient vectors, and we normalize by
\begin{equation}
    N_{b_\text{dist}} =
    \sum_{\vec{k} \in \Omega \cap \Psi} k_\text{gauss} (\vec{k})
\end{equation}
The addition of $b_\text{dist}$ biases matching of features in free-space to other features at a similar distance to surfaces.

Secondly, we augment the descriptor with the type of curvature at each keypoint. In particular, we compute
\begin{equation}
    b_{\text{class}} = \sum_{i=1}^3 [e_i > 0]
\end{equation}
where $e_i$ is the Eigenvalue associated with the Eigenvector $\vec{v}_i$ (calculated in Sec.~\ref{sec:local_reference_frame}), and $[\cdot]$ returns 1 if the condition is true and 0 otherwise. The addition of $b_\text{class}$ to the descriptor biases matching towards features of the same stationary point type. For example, keypoints at maxima of the distance function are biased towards matching other maxima.

We concatenate the $\alpha_\text{dist} b_\text{dist}$ and $\alpha_\text{class} b_\text{class}$ to a flattened (one dimensional) view of the gradient histogram, where $\alpha_\text{dist}$ and $\alpha_\text{class}$ are weighting constants and take the values $1e^{-7}$ and $1e^{-5}$ for the remainder of this work. The final descriptor vector has size $2\times n_{div}^2 + 2$.

\subsection{Registration}
\label{sec:registration}
We treat the place recognition problem as one of pairwise matching. Given two \submap/s, $(\submapmath{S}_i, \submapmath{S}_j)$, representing a putative match, we aim to determine if they are indeed a match, as well as the transform relating their poses $T_{S_iS_j}$. We implement a \ac{RANSAC}-based registration pipeline for this purpose. First, we generate correspondences by performing a K-Nearest-Neighbour search for all keypoints in the query submap, within the target \submap/ (we choose $K=5$ neighbours for the remainder of this work). We accumulate all correspondences in the set $\mathcal{C} = \{c_1, c_2, \dots, c_N\}$, where $c_i$ is a pair of corresponding keypoints i.e. $c_i = \{\vec{p}^k_{S_i}, \vec{p}^l_{S_j}\}$, with $\vec{p}^k_{S_i} \in \mathbb{R}^3$ is the location of the $k^\text{th}$ keypoint in \submap/ $i$.



We select $M=3$ correspondences at random, resulting in a correspondence sub-set, $\mathcal{C}_\text{sub} = \{c_i, c_j, c_k\} \subset \mathcal{C}$. We test this set for geometric consistency by checking that the distances between correspondence endpoints are similar in query and target submaps, a technique first suggested in~\cite{chen20073d} and implemented in Open3D~\cite{Zhou2018}. In particular, we reject correspondence sub-sets not meeting
\begin{equation}
    k_{consist} d_t(c_i, c_j) < d_q(c_i, c_j) < \frac{1}{k_{consist}} d_t(c_i, c_j)
\end{equation}
for all combinations of correspondences pairs $\{c_i, c_j\} \in \mathcal{C}_\text{sub}$, where $d_t(\cdot, \cdot)$ measures the distances between the correspondence endpoints in the target submap, and $d_q$ is similarly defined. We set $k_{consist} = 0.9$ for the remainder of this work. Correspondence sets passing the consistency check are used to estimate transformation candidates $\mat{T}_{S_i S_j}$. 
We rank these candidates based on inlier count, 
\begin{equation}
    score(\mat{T}_{S_i S_j})
    =
    \sum_{c_i \in \mathcal{C}} [d(c_i, \mat{T}_{S_i S_j}) < k_{\text{dist}}]
    \label{eq:inlier_score}
\end{equation}
where
\begin{equation}
    d(c_i, \mat{T}_{S_i S_j})
    =
    \left\lVert \vec{p}^k_{S_i} - \mat{T}_{S_i S_j} \vec{p}^l_{S_j} \right\rVert_2
\end{equation}
where $\mathcal{C}$ denotes a sum over the correspondence set, and $\vec{p}^i_{S_k}$ and $\vec{p}^j_{S_l}$ are the positions of keypoints $i$ and $j$ respectively, and $[\cdot]$ returns 1 if the condition is true and 0 otherwise. Note that $k_\text{dist}$ is dependent on the scale of the environment and is set to various values in our evaluations.

\subsection{\ac{SDF}-based Fitness Evaluation}
\label{sec:validation}

Following \ac{RANSAC}, we calculate a fitness measure for the transformation candidate, $\mat{T}_{S_i S_j}^*$, achieving the highest inlier score, eq.~(\ref{eq:inlier_score}). Global localization decisions are made on the basis of this fitness measure, which we found to lead to improvements in performance, while introducing negligible additional run-time cost.

To calculate the fitness of transformation $\mat{T}_{S_i S_j}^*$, we transform isosurface points from \submap/ $j$, $\mathcal{P}_{S_j} = \{\vec{p}_{\text{iso}, S_j}^i\}_{i=1}^N$, into \submap/ $i$ and calculate a weighted average of the \ac{SDF} value at these points. This operation is performed bidirectionally, i.e. we also transform iso-surface points in \submap/ $i$ into \submap/ $j$. Given a complete, perfectly reconstructed distance function, and the true relative transformation, this score is zero, as all iso-surface points of one \submap/ lie on the \ac{SDF} zero-level set of the other. The expectation in a more realistic setting is that the closer this measure is to zero, the more accurate the transform $\mat{T}_{S_i S_j}^*$. In particular, the fitness cost for $\mat{T}_{S_i S_j}^*$ is given by,
\begin{equation}
    f(\mat{T}_{S_i S_j}^*)
    =
    -\frac{1}{N}
    \left(
    d_{\text{iso},S_i}(\mathcal{P}_{S_j}, \mat{T}_{S_i S_j}^*) + 
    d_{\text{iso},S_j}(\mathcal{P}_{S_i}, \mat{T}_{S_i S_j}^{*-1})
    \right),
    \label{eq:fitness}
\end{equation}
where $d_{\text{iso},S_1}$ is the weighted sum of the \ac{SDF} values,
\begin{equation}
    d_{\text{iso}, S_i}(\mathcal{P}_{S_j}, \mat{T}_{S_i S_j}^*)
    =
    \sum_{\vec{p}_{\text{iso}} \in \mathcal{P}_{S_j}}
    \omega_{S_i}(\mat{T}^*_{S_i S_j} \vec{p}_{\text{iso}})
    \Phi_{S_i}(\mat{T}^*_{S_i S_j} \vec{p}_{\text{iso}}),
\end{equation}
and $d_{\text{iso},S_j}$ is defined similarly. We normalize by the sum of all weights such that,
\begin{equation}
    N =
    \sum_{\vec{p}_\text{iso} \in \mathcal{P}_{S_j}} \omega_{S_i}(\mat{T}^*_{S_i S_j} \vec{p}_{\text{iso}})
    +
    \sum_{\vec{p}_\text{iso} \in \mathcal{P}_{S_i}} \omega_{S_i}(\mat{T}^*_{S_j S_i} \vec{p}_{\text{iso}}).
\end{equation}
The \ac{SDF} function $\Phi$ and $\omega$ are defined sparsely and thus evaluation is not generally possible at all $\vec{p}_\text{iso}$. We require a minimum fraction of points in $\mathcal{P}_{S_i}$ and $\mathcal{P}_{S_j}$ to return valid distances, otherwise the match is not considered. We set this overlap as $k_{\text{overlap}}=15\%$ for all evaluations in this work. Note that the fitness score (\ref{eq:fitness}) has units of of metric length.

\section{Results}
\label{sec:results}

\begin{figure}[t]
    \centering
    \begin{subfigure}[b]{0.8\columnwidth}
    \includegraphics[trim={5cm 0cm 0cm 0cm},clip,width=\columnwidth]{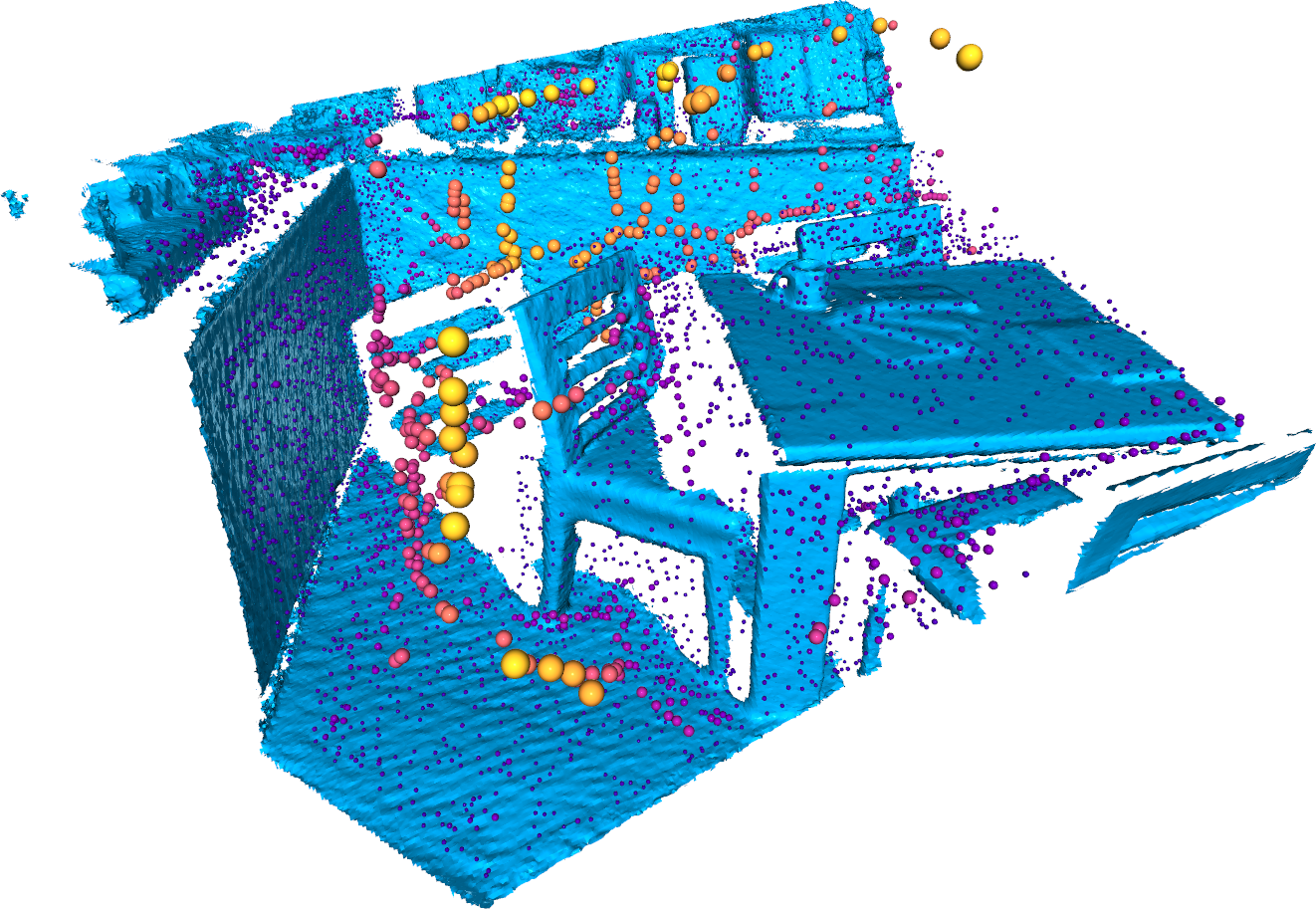}
    \end{subfigure}
    \begin{subfigure}[b]{0.05\columnwidth}
        \input{figures/3dmatch/color_bar.pgf}
    \end{subfigure}
    \caption{A scene fragment from the 3DMatch dataset~\cite{zeng20173dmatch}, analyzed in Sec.~\ref{sec:results_3dmatch}. Colored spheres show the keypoint locations described by our method. The size and color of the keypoints are modulated by the distance from the keypoint to the closest surface. The plot shows features located in free-space; these are particularly visible in the foreground, between the chair and the cabinet.}
    \label{fig:3dmatch_features}
    \vspace{-4mm}
\end{figure}

\begin{figure}
    \centering
    \begin{subfigure}[b]{0.8\columnwidth}
    \includegraphics[trim={5cm -7.5cm 0cm 0cm},clip,width=\columnwidth]{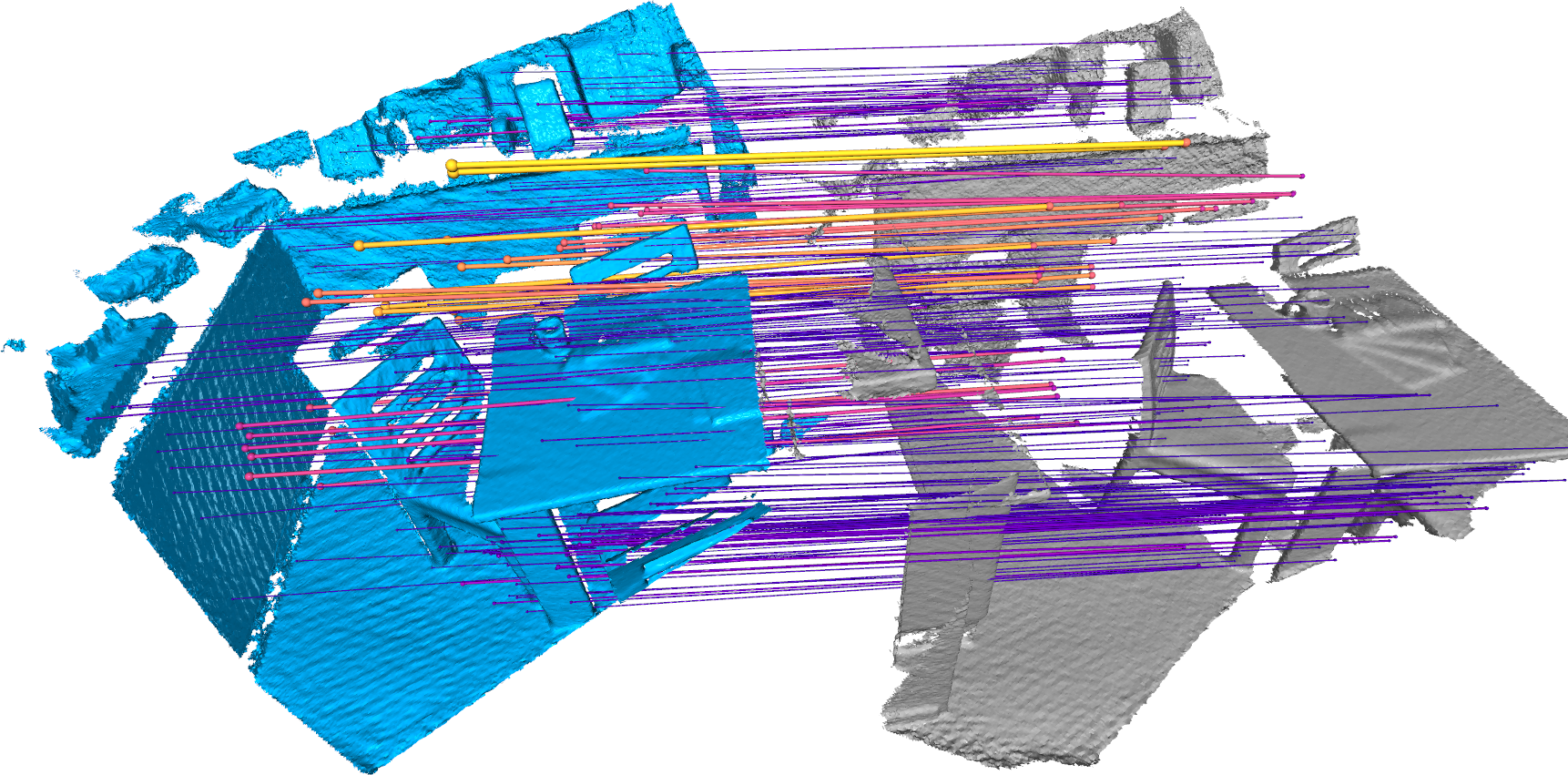}
    \end{subfigure}
    \begin{subfigure}[b]{0.05\columnwidth}
        \input{figures/3dmatch/color_bar.pgf}
    \end{subfigure}
    \vspace{-4mm}
    \caption{An example match between two fragments in the 3DMatch dataset~\cite{zeng20173dmatch}. We show a random subset of inlier features and correspondences that generated the match. The size and color of the keypoints and correspondence lines are modulated by the distance from the keypoint(s) to the closest surface. This alignment resulted in \texttildelow1200 inliers.}
    \label{fig:subsampling_reconstruction}
    \vspace{-4mm}
\end{figure}

%
\begin{table*}
\parbox[t][][t]{1.1\columnwidth}{
\centering
\begin{tabular}{@{}rrrrrrrrrr@{}}
\toprule
\phantom{abc} & \multicolumn{8}{c}{Recall at Precision 0.8} \\
\cmidrule{2-10}
\phantom{abc} & \multicolumn{8}{c}{3DMatch Validation Dataset Index} \\
Method & \#1 & \#2 & \#3 & \#4 & \#5 & \#6 & \#7 & \#8 & Mean\\
\midrule
fpfh (de.) & 0.46 & 0.21 & 0.10 & 0.12 & 0.14 & 0.20 & 0.31 & 0.12 & 0.21 \\ 
shot (de.) & 0.54 & 0.23 & 0.15 & 0.13 & 0.14 & 0.22 & 0.34 & 0.11 & 0.23 \\ 
fpfh (sp.) & 0.46 & 0.34 & 0.13 & 0.20 & 0.19 & 0.35 & 0.33 & 0.11 & 0.26 \\ 
shot (sp.) & \textit{0.57} & \textit{0.39} & \textit{0.16} & \textit{0.23} & \textit{0.25} & \textbf{0.44} & \textit{0.35} & \textit{0.14} & \textit{0.32} \\ 
freespace & \textbf{0.70} & \textbf{0.44} & \textbf{0.20} & \textbf{0.28} & \textbf{0.25} & \textit{0.43} & \textbf{0.36} & \textbf{0.16} & \textbf{0.35} \\ 
\midrule
Rel. (\%) & \textbf{21.8} & \textbf{12.1} & \textbf{20.9} & \textbf{20.4} & \textbf{2.8} & -1.8 & \textbf{2.3} & \textbf{13.9} & \textbf{11.5} \\ 
\bottomrule
\end{tabular}
\caption{Results of pairwise matching on the 3DMatch dataset~\cite{zeng20173dmatch}. Two baseline surface-based features, \ac{FPFH} and SHOT, with two different \ac{RANSAC} validation methods, correspondence-based (sp) and pointcloud-based (de). The final line shows percentage performance difference between the proposed method and the best surface descriptor. Best performance is in bold text, the second best in italics.}
\label{tab:results_3dmatch}
} 
\hfill
\parbox[t][][t]{0.9\columnwidth}{
\centering
\begin{tabular}{@{}rrrrrrrr@{}}
\toprule
\phantom{abc} & \multicolumn{7}{c}{Recall at Precision 0.8} \\
\cmidrule{2-8}
\phantom{abc} & \multicolumn{7}{c}{Deutsches Museum Dataset Index} \\
Method & \#1 & \#2 & \#3 & \#4 & \#5 & \#6 & Mean\\
\midrule
fpfh (vox) & 0.31 & 0.57 & 0.39 & 0.42 & 0.36 & 0.41 & 0.41 \\ 
fpfh (iso) & 0.34 & 0.61 & 0.44 & \textit{0.48} & 0.44 & 0.46 & 0.46 \\ 
shot (vox) & 0.21 & 0.55 & 0.40 & 0.44 & 0.39 & 0.41 & 0.40 \\ 
shot (iso) & \textit{0.39} & \textit{0.61} & \textit{0.48} & 0.46 & \textit{0.47} & \textit{0.48} & \textit{0.48} \\ 
freespace & \textbf{0.58} & \textbf{0.68} & \textbf{0.51} & \textbf{0.50} & \textbf{0.56} & \textbf{0.56} & \textbf{0.56} \\ 
\midrule
Rel. (\%) & \textbf{49.5} & \textbf{10.9} & \textbf{7.4} & \textbf{4.3} & \textbf{19.3} & \textbf{15.5} & \textbf{17.8} \\
\bottomrule
\end{tabular}
\caption{Results of the Deutsches museum experiment described in Sec.~\ref{sec:results_deutsches}. The table shows recall values generated by each method at 0.8 precision. The final line shows percentage performance difference between the proposed method and the best surface descriptor. Best performance is in bold text, the second best in italics.}
\label{tab:results_deutsches_museum}
} 
\vspace{-5mm}
\end{table*}

In this section, we aim to validate the hypotheses of this paper, that a) the proposed system enables global place recognition in a collection of \ac{SDF} \submap/s by extracting features \textit{directly} on the \ac{SDF}; that b) the use of these features can increase localization performance when compared to existing features computed based on surface information only; and that c) the use of free-space information is integral to the improvements in performance we show.

\subsection{Parameters}

The performance of both our proposal and the baseline methods are affected by parameters. We found our proposal to be robust across the various environments in this section and use fixed parameter values as described in the method in all evaluations. We set the feature radius $r_f=15\,\text{voxels}$, and number of angular divisions $n_\text{div}=10\,\text{divs}/180^{\circ}$. For \ac{FPFH} and \ac{SHOT} in the 3DMatch~\cite{zeng20173dmatch} dataset, we use feature radius and normal calculation radius values provided by the authors in~\cite{gojcic2019perfect}, where a similar analysis is performed. For the Deutsches Museum dataset we grid searched for optimal parameters, resulting in feature radii of 1.5\,m and 3.0\,m for \ac{FPFH} and \ac{SHOT} respectively. We selected the scale-dependant \ac{RANSAC} inlier threshold, $k_\text{dist}$, for all methods through a grid search for values leading to optimal results.

\subsection{3DMatch Dataset}
\label{sec:results_3dmatch}

\begin{figure}
    \centering
    \setlength{\figurewidth}{0.75\columnwidth}
    \setlength{\figureheight}{0.4\columnwidth}
    \input{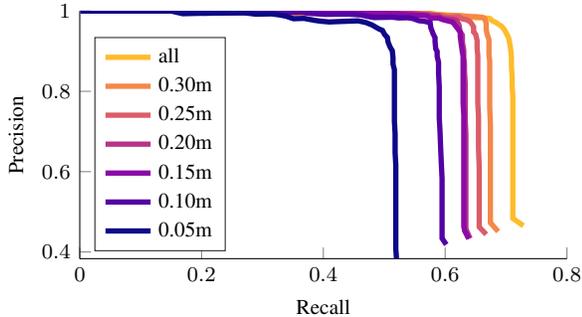}
    \caption{Precision-recall curves generated by eliminating features in free-space to varying degrees (Sec.~\ref{sec:results_3dmatch}). The curve for 0.05\,m for example only permits feature extraction within 5\,cm of a surface. The figure shows that features in free-space contribute significantly to the proposed method's performance.}
    \label{fig:results_3dmatch_distance_elimination}
    \vspace{-4mm}
\end{figure}

We quantify the performance of the proposed method for pairwise alignment of scene fragments using the 3DMatch dataset~\cite{zeng20173dmatch}. This dataset consists of reconstructions of 8 indoor scenes scanned with an \rgbd/ sensor. We use the provided toolbox to fuse sequences of consecutive \rgbd/ images to form partially overlapping surface pointclouds. Additionally, we generate \ac{SDF}s using cblox~\cite{millane2018c}, which form the input to our system. 
We follow the approach in~\cite{zeng20173dmatch} and sample 5000 random surface points as keypoints for the surface-based descriptors. To control for feature number, we also limit our proposed method to extracting the 5000 keypoints, chosen to have the strongest \ac{DoH} response, eq. \eqref{eq:doh_response}. An example scene fragment and extracted features are shown in Fig.~\ref{fig:3dmatch_features}. To determine ground truth match/non-match labels, we calculate the overlapping volume between each pair of voxel grids. Pairs sharing more than $1\,\text{m}^3$ overlap are considered ground-truth matches. Estimated labels are computed by thresholding fitness values from registration. We generate precision-recall curves by varying this fitness threshold. For pairs testing positive for a match, we additionally require the relative fragment pose to be within 0.2\,m of the true value. 

We compare the proposed approach to \ac{FPFH}~\cite{rusu2009fast} and \ac{SHOT}~\cite{salti2014shot} features, which have shown state-of-the-art performance in comparisons of handcrafted features~\cite{gawel20173d}, and have open-source implementations released with PCL~\cite{rusu20113d}. For both features, we test two \ac{RANSAC}-based registration pipelines. The first was proposed with \ac{FPFH} in~\cite{rusu2009fast}. This approach uses a pointcloud-based fitness measure, and is implemented as part of Open3D~\cite{Zhou2018}. The second approach uses a fitness measure based on keypoint distance, analogous to our $score$ (\ref{eq:inlier_score}). We denote these methods \textit{dense} and \textit{sparse} respectively in Table~\ref{tab:results_3dmatch}. Note that we perform the geometric consistency checks described in Sec.~\ref{sec:registration} for all methods. We set the number of \ac{RANSAC} iterations to $4e^6$ for all experiments.

Table~\ref{tab:results_3dmatch} shows the results of this experiment, summarized as recall rates at precision 0.8. Some generalizations are possible. \ac{SHOT} outperformed \ac{FPFH}, which agrees with similar analyses~\cite{gawel20173d, gojcic2019perfect}, and reflects \ac{SHOT}s greater dimensionality: 352 vs. 33. Registration using the keypoint-based fitness measure, rather than the method proposed in~\cite{rusu2009fast}, led to better results for both feature types. Note that the approach of selecting (random) keypoints and computing correspondences for these keypoints-only, also avoids dense correspondence computation. The proposed \ac{SDF}-based feature outperformed the competing methods, leading to an average improvement in recall rates of $11.5\%$ with respect to the runner-up method, \ac{SHOT} (sparse). The proposed descriptor has dimension 202, lying between the dimensionality of the baseline methods.

Finally, we analyze the contribution of features in free-space to the performance of the proposed method. We generate precision-recall curves, as before, however we limit the proposed method to extracting features near surface boundaries, to varying degrees. In particular, we restrict feature extraction such that each keypoint lies at a distance from surfaces less than a threshold, $d_\text{lim}$ for  $d_\text{lim} \in \{0.30, 0.25, 0.20, 0.15, 0.10, 0.05\}\,\text{m}$. Fig.~\ref{fig:results_3dmatch_distance_elimination} results of this study for 3DMatch dataset \#1. The curves show that the use of features in free-space generates significant performance gains, validating one of the hypotheses of this paper.



\subsection{Deutsches Museum}
\label{sec:results_deutsches}

\begin{figure}
    \centering
    \includegraphics[width=0.95\columnwidth,trim=0 0 0 0]{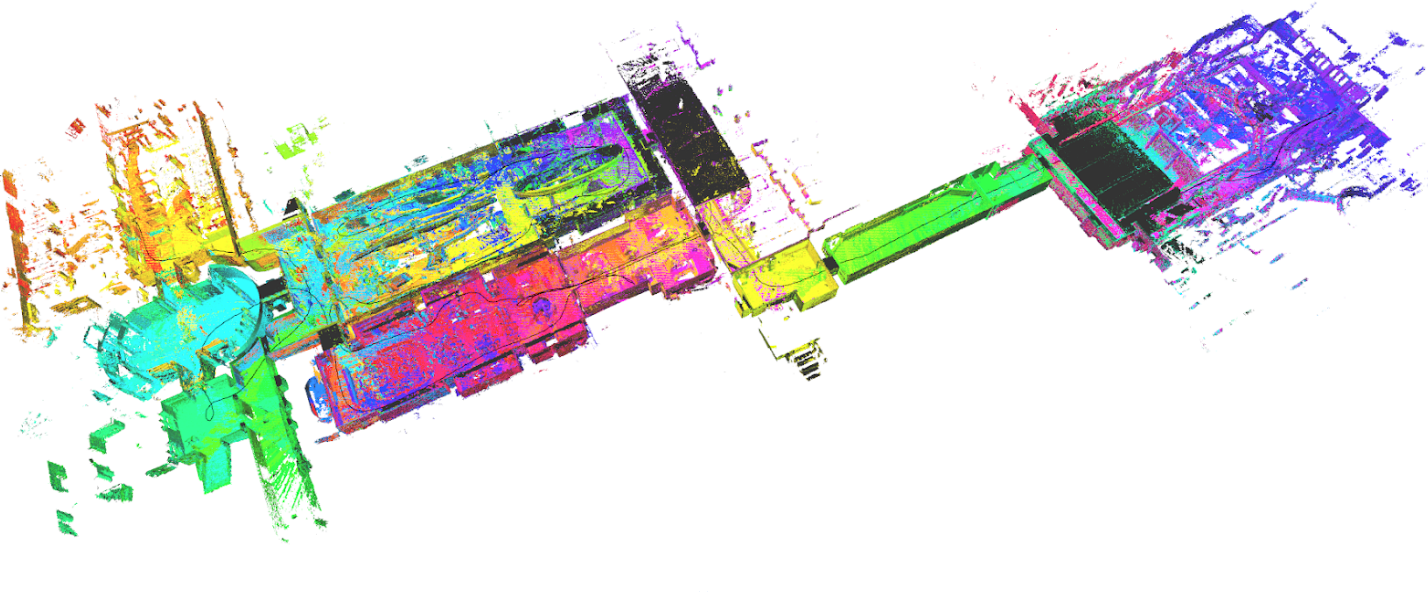}
    \caption{A mesh extracted from an \ac{SDF} map of part of the Deutsches museum (Sec.~\ref{sec:results_deutsches}). The reconstruction is performed using Cartographer~\cite{hess2016real} and cblox~\cite{millane2018c}. The color of the mesh at each location indicates the \submap/ contributing to the reconstruction at that location.}
    \label{fig:results_deutsches_museum_map}
    \vspace{-4mm}
\end{figure}


In order to evaluate our system for use with 3D LiDAR data we perform evaluations on the Deutsches Museum dataset, released as part of Google’s Cartographer SLAM system~\cite{hess2016real}. This dataset provides sensor data from backpack mounted 3D LiDARs and an \ac{IMU}, and was collected during walking traversals through a museum. We select 6 trajectories of varying sizes with which to perform evaluation. The trajectories range in length between \texttildelow5\,minutes/35\,\submap/s and \texttildelow20\,minutes/136\,\submap/s.

\begin{figure*}[t]
    \centering
    \includegraphics[width=0.80\textwidth,trim=0 0 0 0]{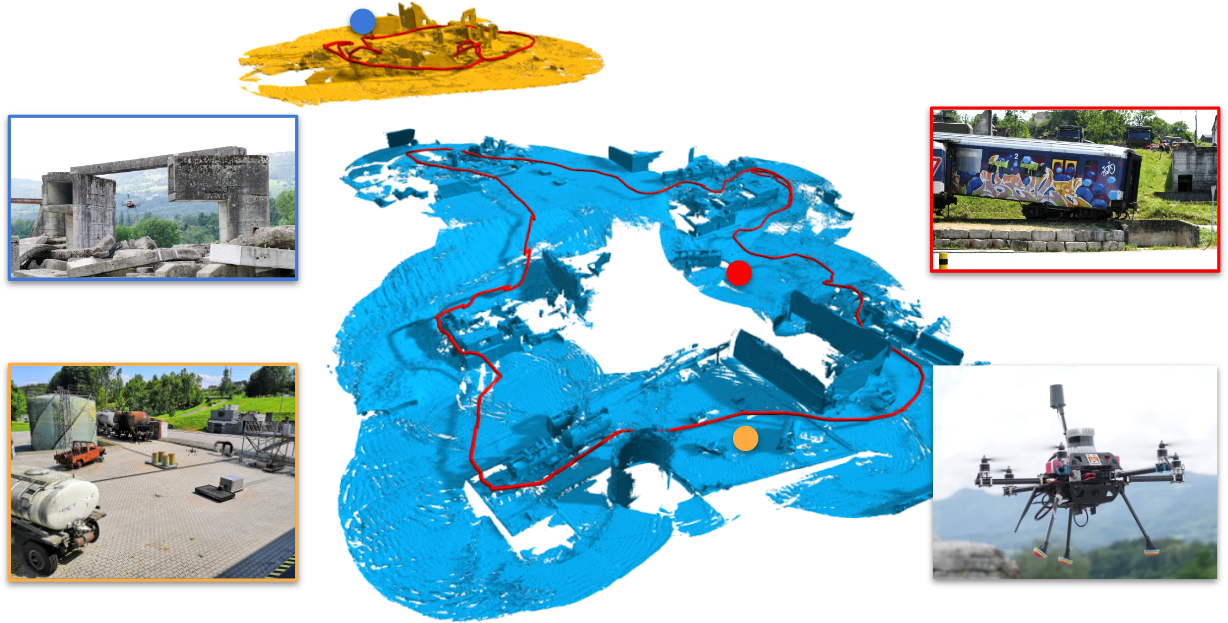}
    \caption{Two maps used for the localization experiment described in Sec.~\ref{sec:results_arche}. The maps are reconstructed using using our previous work voxgraph~\cite{reijgwart2019voxgraph} and data collected during two flights of an \ac{MAV} equipped with a monocular camera, inertial sensing, a 3D LiDAR (visible in the lower-right of the figure). The maps partially overlap at the ruins in the top left of the figure. A successful localization is shown in Figs.~\ref{fig:results_arche_match} and \ref{fig:teaser}.}
    \label{fig:results_arche_maps}
\vspace{-5mm}
\end{figure*}

To produce a collection of \ac{SDF} \submap/s with which to evaluate our system, we compute a globally optimized trajectory using Cartographer~\cite{hess2016real} and accumulate LiDAR data in \submap/ \acp{TSDF} using cblox~\cite{millane2018c}. Fig.~\ref{fig:results_deutsches_museum_map} shows an example map, where varying surface color indicates the change in the \submap/ contributing to the reconstruction at each location. To produce pointcloud \submap/s for surface descriptors we follow two approaches. In the first approach, we fuse sensor pointclouds using voxel filtering. In the second, we extract iso-surface pointclouds from the \ac{SDF} \submap/s described above. The result of these processes are collections of co-located \submap/s, computed from the same sensor data, represented as both \ac{SDF}s and pointclouds. We generate groundtruth and estimated match/non-match labels using \submap/ overlap and fitness thresholds as described in Sec.~\ref{sec:results_3dmatch}.

We compare our approach against \ac{FPFH} and \ac{SHOT} as in Sec.~\ref{sec:results_3dmatch}. For \ac{FPFH} we follow the approach proposed in~\cite{rusu2009fast}; we extract features densely and use a pointcloud-based registration pipeline to avoid computing dense correspondences. For \ac{SHOT} we follow the approach which lead to the best results in Sec.~\ref{sec:results_3dmatch}: we randomly select 5000 surface points as keypoints for feature extraction and use a keypoint-based fitness measure. Again, we also limit our proposed method to extracting at most 5000 keypoints.

Table~\ref{tab:results_deutsches_museum} summarizes the results of this experiment as recall values at precision 0.8. Some generalizations can be made. For the baseline features, iso-surface pointcloud extraction produced better results than voxel filtering, leading to mean improvements of 12.6\% for \ac{FPFH}, and 26.7\% for \ac{SHOT}. This is a somewhat reassuring result. To generate an \ac{SDF} iso-surface, pointcloud data is first fused into a voxelized \ac{TSDF}, which one might assume to introduce some discretization error when compared with the voxel filtering approach. These results show that, in this case at least, \ac{TSDF} fusion has a positive effect on performance. Our proposed feature outperforms the baseline methods. With respect to the runner-up method, which was SHOT (iso-surface) is all but one experiment (in which it was \ac{FPFH} (iso-surface)), our proposal generates a mean improvement of 17.8\%. 


\subsection{\ac{MAV} Dataset}
\label{sec:results_arche}

To demonstrate the utility and generalization of our system for robotic applications, we demonstrate its use for localizing an \acf{MAV} against an existing map. We recorded LiDAR, camera and inertial measurements from sensors attached to an \ac{MAV} (Fig.~\ref{fig:results_arche_maps}) flying through a search and rescue training area. The target trajectory was \texttildelow500\,m in length (map area: \texttildelow120\,m$\times$120\,m), and the query trajectory \texttildelow200\,m (map area: \texttildelow70\,m$\times$70\,m). We performed visual-inertial odometry using ROVIO~\cite{bloesch2017iterated} for local motion tracking and built a globally consistent \submap/-based map using our previous work voxgraph~\cite{reijgwart2019voxgraph}.  
Fig.~\ref{fig:results_arche_maps} shows the two maps positioned relative to one another using \ac{RTK}-GPS (with the query map raised in the height for clarity). The section of overlap occurs at the ruins of a badly damaged building, see Fig.~\ref{fig:results_arche_maps} (top left) for a photo. We use the RTK-GPS system to determine ground-truth \submap/ poses and generate ground-truth match/non-match labels.

\begin{figure}
    \centering
    \includegraphics[width=0.90\columnwidth]{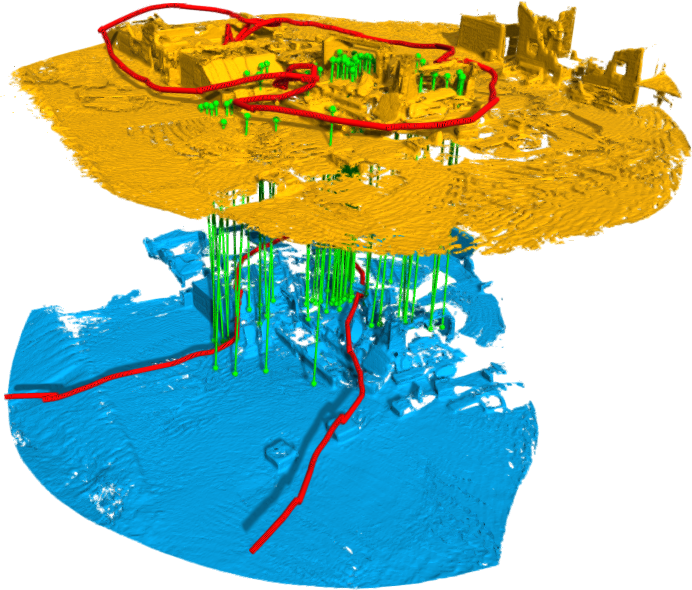}
    \caption{A successful \submap/-\submap/ match between the query map (yellow) and the target map (blue). The query map is shown elevated with respect to target map. Inlier features leading to the match are shown in green with vertical bars connecting corresponding features. The final alignment, following \ac{ICP}, is shown in Fig.~\ref{fig:teaser}}
    \label{fig:results_arche_match}
    \vspace{-5mm}
\end{figure}

We perform localization by exhaustively matching each query \submap/ against all \submap/s in the target collection and calculate estimated labels. We achieve a recall rate of 0.54 at precision 0.8 indicating the applicability of the proposed system to a new environment, with only a single parameter change $k_\text{dist}$. Fig.~\ref{fig:results_arche_match} shows the results of one successful match linked by the inlier features that produced it, and Fig.~\ref{fig:teaser} the final registration.

\section{Conclusion}
\label{sec:conclusion}

In this paper we have presented a novel approach for localization in maps represented as collections of \ac{SDF} \submap/s. Our system extracts features directly on the \ac{SDF}, allowing description of both surface and free-space geometry. We suggest an interest point detector based on extrema of the \ac{DoH} volume, which selects for regions of high curvature in the distance field. Keypoints are described using a spherical gradient histogram, augmented with two pieces of \ac{SDF} specific information: feature distance, and curvature class. We test our approach on a variety of environments and sensors. In the \rgbd/ camera-based 3DMatch dataset~\cite{zeng20173dmatch}, we show a mean improvement of \texttildelow12\% over the best baseline method. In the LiDAR-based Cartographer dataset~\cite{hess2016real} we show a mean improvement of \texttildelow18\% over the closest baseline method. Finally, we demonstrate our method for localizing an \ac{MAV} against an earlier flight through a search and rescue training ground. Our proposal requires only changing a single scale-dependant parameter when applied to these three different environments and sensing setups, an indication of its generality. Future work will investigate if our approach, extracting features on the \ac{SDF}, is advantageous in the domain of deep-learned descriptors that have become popular in recent years.

\bibliographystyle{ieeetr}
\bibliography{bibliography}

\begin{thebibliography}{10}

\bibitem{cadena2016past}
C.~Cadena, L.~Carlone, H.~Carrillo, Y.~Latif, D.~Scaramuzza, J.~Neira, I.~Reid,
  and J.~J. Leonard, ``Past, present, and future of simultaneous localization
  and mapping: Toward the robust-perception age,'' {\em TRO}, vol.~32, no.~6,
  pp.~1309--1332, 2016.

\bibitem{bosse2013place}
M.~Bosse and R.~Zlot, ``Place recognition using keypoint voting in large 3d
  lidar datasets,'' in {\em ICRA}, pp.~2677--2684, 2013.

\bibitem{steder2011place}
B.~Steder, M.~Ruhnke, S.~Grzonka, and W.~Burgard, ``Place recognition in 3d
  scans using a combination of bag of words and point feature based relative
  pose estimation,'' in {\em IROS}, pp.~1249--1255, 2011.

\bibitem{dube2020segmap}
R.~Dub{\'e}, A.~Cramariuc, D.~Dugas, H.~Sommer, M.~Dymczyk, J.~Nieto,
  R.~Siegwart, and C.~Cadena, ``Segmap: Segment-based mapping and localization
  using data-driven descriptors,'' {\em IJRR}, vol.~39, no.~2-3, pp.~339--355,
  2020.

\bibitem{rusu2009fast}
R.~B. Rusu, N.~Blodow, and M.~Beetz, ``Fast point feature histograms (fpfh) for
  3d registration,'' in {\em ICRA}, pp.~3212--3217, 2009.

\bibitem{choi2015robust}
S.~Choi, Q.-Y. Zhou, and V.~Koltun, ``Robust reconstruction of indoor scenes,''
  in {\em CVPR}, pp.~5556--5565, 2015.

\bibitem{salti2014shot}
S.~Salti, F.~Tombari, and L.~Di~Stefano, ``Shot: Unique signatures of
  histograms for surface and texture description,'' {\em CVIU}, vol.~125,
  pp.~251--264, 2014.

\bibitem{reijgwart2019voxgraph}
V.~Reijgwart, A.~Millane, H.~Oleynikova, R.~Siegwart, C.~Cadena, and J.~Nieto,
  ``Voxgraph: Globally consistent, volumetric mapping using signed distance
  function submaps,'' {\em RA-L}, vol.~5, no.~1, pp.~227--234, 2019.

\bibitem{fioraio2015large}
N.~Fioraio, J.~Taylor, A.~Fitzgibbon, L.~Di~Stefano, and S.~Izadi,
  ``Large-scale and drift-free surface reconstruction using online subvolume
  registration,'' in {\em CVPR}, pp.~4475--4483, 2015.

\bibitem{oleynikova2017voxblox}
H.~Oleynikova, Z.~Taylor, M.~Fehr, R.~Siegwart, and J.~Nieto, ``Voxblox:
  Incremental 3d euclidean signed distance fields for on-board mav planning,''
  in {\em IROS}, pp.~1366--1373, 2017.

\bibitem{millane2019freespace}
A.~{Millane}, H.~{Oleynikova}, J.~{Nieto}, R.~{Siegwart}, and C.~{Cadena},
  ``Free-space features: Global localization in 2d laser slam using distance
  function maps,'' in {\em IROS}, pp.~1271--1277, 2019.

\bibitem{izadi2011kinectfusion}
S.~Izadi, D.~Kim, O.~Hilliges, D.~Molyneaux, R.~Newcombe, P.~Kohli, J.~Shotton,
  S.~Hodges, D.~Freeman, A.~Davison, {\em et~al.}, ``Kinectfusion: real-time 3d
  reconstruction and interaction using a moving depth camera,'' in {\em ACM
  symposium UIST}, pp.~559--568, 2011.

\bibitem{niessner2013real}
M.~Nie{\ss}ner, M.~Zollh{\"o}fer, S.~Izadi, and M.~Stamminger, ``Real-time 3d
  reconstruction at scale using voxel hashing,'' {\em ACM ToG}, vol.~32, no.~6,
  pp.~1--11, 2013.

\bibitem{dai2017bundlefusion}
A.~Dai, M.~Nie{\ss}ner, M.~Zollh{\"o}fer, S.~Izadi, and C.~Theobalt,
  ``Bundlefusion: Real-time globally consistent 3d reconstruction using
  on-the-fly surface reintegration,'' {\em ACM ToG}, vol.~36, no.~4, p.~1,
  2017.

\bibitem{millane2018c}
A.~Millane, Z.~Taylor, H.~Oleynikova, J.~Nieto, R.~Siegwart, and C.~Cadena,
  ``C-blox: A scalable and consistent tsdf-based dense mapping approach,'' in
  {\em IROS}, pp.~995--1002, 2018.

\bibitem{lowe1999object}
D.~G. Lowe, ``Object recognition from local scale-invariant features,'' in {\em
  ICCV}, vol.~2, pp.~1150--1157, Ieee, 1999.

\bibitem{glocker2014real}
B.~Glocker, J.~Shotton, A.~Criminisi, and S.~Izadi, ``Real-time rgb-d camera
  relocalization via randomized ferns for keyframe encoding,'' {\em TVCG},
  vol.~21, no.~5, pp.~571--583, 2014.

\bibitem{lowry2015visual}
S.~Lowry, N.~S{\"u}nderhauf, P.~Newman, J.~J. Leonard, D.~Cox, P.~Corke, and
  M.~J. Milford, ``Visual place recognition: A survey,'' {\em TRO}, vol.~32,
  no.~1, pp.~1--19, 2015.

\bibitem{behley2018rss}
J.~Behley and C.~Stachniss, ``Efficient surfel-based slam using 3d laser range
  data in urban environments,'' in {\em RSS}, 2018.

\bibitem{hess2016real}
W.~Hess, D.~Kohler, H.~Rapp, and D.~Andor, ``Real-time loop closure in 2d lidar
  slam,'' in {\em ICRA}, pp.~1271--1278, 2016.

\bibitem{gawel20173d}
A.~Gawel, R.~Dub{\'e}, H.~Surmann, J.~Nieto, R.~Siegwart, and C.~Cadena, ``3d
  registration of aerial and ground robots for disaster response: An evaluation
  of features, descriptors, and transformation estimation,'' in {\em SSRR},
  pp.~27--34, 2017.

\bibitem{petrelli2011repeatability}
A.~Petrelli and L.~Di~Stefano, ``On the repeatability of the local reference
  frame for partial shape matching,'' in {\em ICCV}, pp.~2244--2251, 2011.

\bibitem{dalal2005histograms}
N.~Dalal and B.~Triggs, ``Histograms of oriented gradients for human
  detection,'' in {\em CVPR}, vol.~1, pp.~886--893, 2005.

\bibitem{scovanner20073}
P.~Scovanner, S.~Ali, and M.~Shah, ``A 3-dimensional sift descriptor and its
  application to action recognition,'' in {\em ACM MM}, pp.~357--360, 2007.

\bibitem{chen20073d}
H.~Chen and B.~Bhanu, ``3d free-form object recognition in range images using
  local surface patches,'' {\em Pattern Recognition Letters}, vol.~28, no.~10,
  pp.~1252--1262, 2007.

\bibitem{Zhou2018}
Q.-Y. Zhou, J.~Park, and V.~Koltun, ``{Open3D}: {A} modern library for {3D}
  data processing,'' {\em arXiv:1801.09847}, 2018.

\bibitem{zeng20173dmatch}
A.~Zeng, S.~Song, M.~Nie{\ss}ner, M.~Fisher, J.~Xiao, and T.~Funkhouser,
  ``3dmatch: Learning local geometric descriptors from rgb-d reconstructions,''
  in {\em CVPR}, 2017.

\bibitem{gojcic2019perfect}
Z.~Gojcic, C.~Zhou, J.~D. Wegner, and A.~Wieser, ``The perfect match: 3d point
  cloud matching with smoothed densities,'' in {\em CVPR}, pp.~5545--5554,
  2019.

\bibitem{rusu20113d}
R.~B. Rusu and S.~Cousins, ``3d is here: Point cloud library (pcl),'' in {\em
  ICRA}, pp.~1--4, 2011.

\bibitem{bloesch2017iterated}
M.~Bloesch, M.~Burri, S.~Omari, M.~Hutter, and R.~Siegwart, ``Iterated extended
  kalman filter based visual-inertial odometry using direct photometric
  feedback,'' {\em IJRR}, vol.~36, no.~10, pp.~1053--1072, 2017.

\end{thebibliography}


\end{document}